\documentclass[10pt,twocolumn,letterpaper]{article}

\usepackage{iccv}
\usepackage{times}
\usepackage{epsfig}
\usepackage{graphicx}
\usepackage{amsmath}
\usepackage{amssymb}
\usepackage{graphicx}
\usepackage{amsmath}
\usepackage{amssymb}
\usepackage{booktabs}
\usepackage{bbm}
\usepackage[table]{xcolor}
\usepackage{colortbl}
\usepackage{multirow}
\usepackage{mmstyles}
\usepackage{bbm}
\usepackage{algorithm}
\usepackage{algpseudocode}
\definecolor{myy}{RGB}{126,95,0}
\definecolor{mygray}{gray}{.9}
\definecolor{mygray2}{gray}{0.5}
\definecolor{bblue}{RGB}{30,80,120}
\definecolor{mygray1}{gray}{.7}
\definecolor{ggray}{RGB}{127,127,127}
\definecolor{mygreen}{RGB}{93,174,86}
\definecolor{darkergreen}{RGB}{21, 152, 56}
\definecolor{red2}{RGB}{252, 54, 65}
\definecolor{y}{RGB}{255,230,0}
\definecolor{p}{RGB}{236,185,255}
\definecolor{g}{RGB}{0,235,0}

\usepackage[pagebackref=true,breaklinks=true,letterpaper=true,colorlinks,bookmarks=false]{hyperref}

\iccvfinalcopy 

\def\iccvPaperID{ } 

\ificcvfinal\fi

\begin{document}

\title{Open-World Weakly-Supervised Object Localization}

\author{Jinheng Xie$^{1\dagger}$, Zhaochuan Luo$^{1\dagger}$, Yuexiang Li$^2$, Haozhe Liu$^3$, Linlin Shen$^{1*}$, Mike Zheng Shou$^4$ \\
$^1$Shenzhen University, 
$^2$Tencent Jarvis Lab,
$^3$KAUST,
$^4$National University of Singapore\\
{\tt\small \{xiejinheng2020,luozhaochuan2021\}@email.szu.edu.cn, llshen@szu.edu.cn}
}

\maketitle


\begin{abstract}
   While remarkable success has been achieved in weakly-supervised object localization (WSOL), current frameworks are not capable of locating objects of novel categories in open-world settings. To address this issue, we are the first to introduce a new weakly-supervised object localization task called \textbf{OWSOL} (Open-World Weakly-Supervised Object Localization). During training, all labeled data comes from known categories and, both known and novel categories exist in the unlabeled data. To handle such data, we propose a novel paradigm of contrastive representation co-learning using both labeled and unlabeled data to generate a complete \textbf{G-CAM} (Generalized Class Activation Map) for object localization, without the requirement of bounding box annotation. As no class label is available for the unlabelled data, we conduct clustering over the full training set and design a novel multiple semantic centroids-driven contrastive loss for representation learning. We re-organize two widely used datasets, i.e., ImageNet-1K and iNatLoc500, and propose OpenImages150 to serve as evaluation benchmarks for OWSOL. Extensive experiments demonstrate that the proposed method can surpass all baselines by a large margin. We believe that this work can shift the close-set localization towards the open-world setting and serve as a foundation for subsequent works. Code will be released at \url{https://github.com/ryylcc/OWSOL}.
\end{abstract}

\let\thefootnote\relax\footnotetext{$\dagger$ Equal Contribution}
\let\thefootnote\relax\footnotetext{* Corresponding Author}

\begin{figure}[t]
	\centering
    \includegraphics[width=\linewidth]{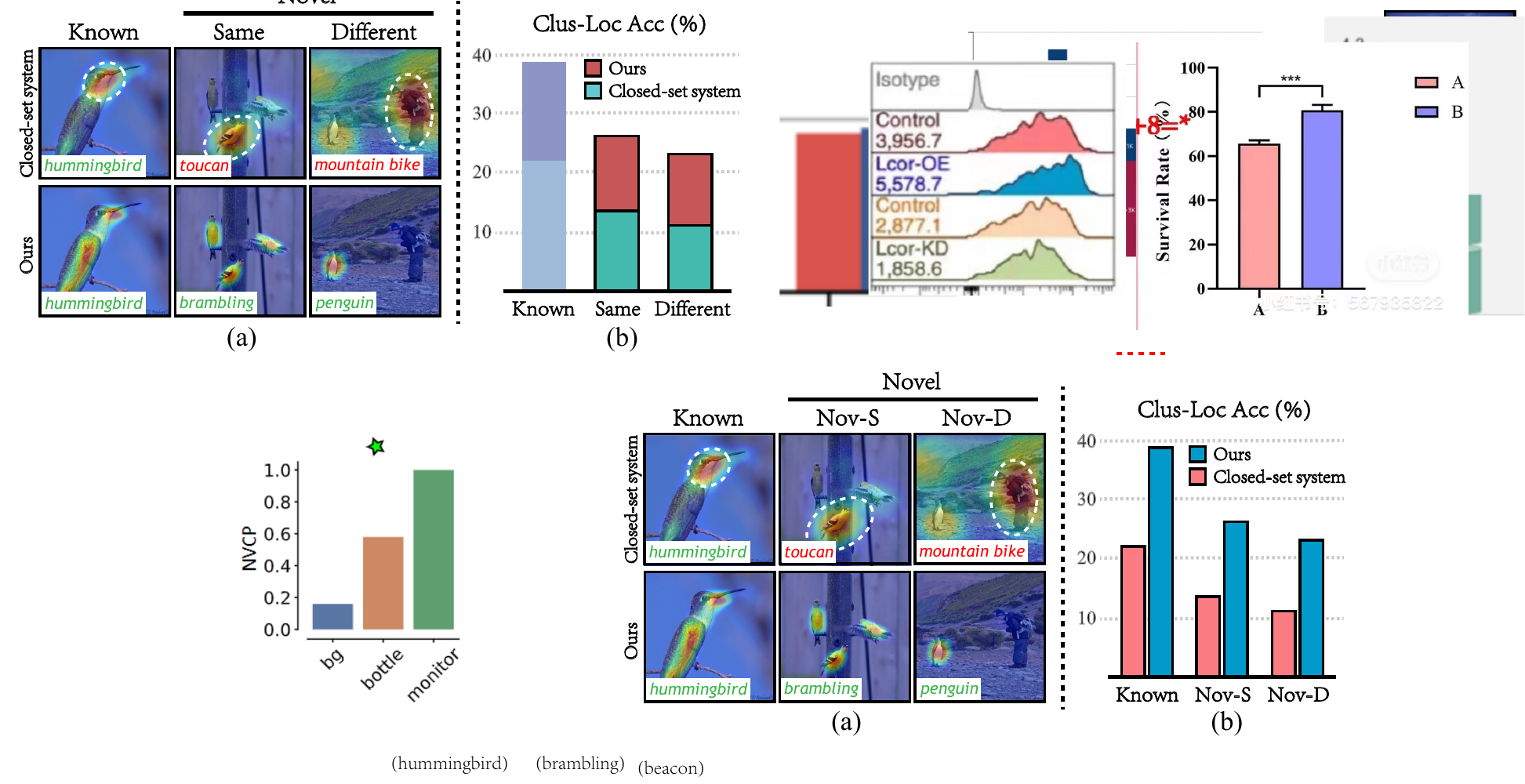}
	\vspace{-15pt}
	\caption{Illustration of the deficiency of a closed-set localization system and comparison with the proposed method under an open-world setting. One can observe from (a) that the closed-set system fails to classify the novel categories and attend to the wrong or only discriminative object regions. On the contrary, the proposed method can recognize known and novel categories with accurate attention. (b) demonstrates that our method can consistently achieve better performance on both known and novel categories.} 
	\label{fig:fig1}
	\vspace{-20pt}
\end{figure}
\section{Introduction}
\label{sec:intro} 
Over the past decade, with massive labeled data, deep neural networks have surpassed other classical machine learning algorithms in visual recognition and detection. However, it requires time-consuming and labor-intensive procedures to generate high-quality labeled data, especially in object detection. To reduce the workload, many kinds of weak supervision, e.g., image-level tags, have been explored in weakly-supervised object localization (WSOL).

In recent years, great efforts and promising results have been derived from WSOL~\cite{ewsol, RCAM, Pan_2021_CVPR, STL, Wei_2021_CVPR, FAM, TS-CAM, kim2021normalization, lctr, Kim_2022_CVPR, OLDA, BAS, CREAM, C2AM}. It significantly reduces the manual cost of annotation but is still far from applicable to the general scene, i.e., open-world situations. Current settings for WSOL assume that all the data comes from a fixed number of known categories, i.e., a closed-world setting. Such an assumption is not valid in the general scene. For example, with a cloud-based object localization system, customers can always upload out-of-distribution images, i.e., \textit{novel categories}. Such a closed-set system, e.g., CAM~\cite{cam}, would potentially fail to classify and localize those novel categories without any defined a-priori. We illustrate the deficiency of a closed-set system in Figure \ref{fig:fig1} (a). As observed, the system can correctly classify the \textbf{Known} category, e.g., 'hummingbird', but the corresponding CAM only localizes the head regions. For the \textbf{Nov}el birds from the \textbf{S}ame family as the known, termed \textbf{Nov-S}, such as 'brambling', the system mistakenly identifies it as `toucan' and the extracted CAM cannot be used to precisely localize the object, due to excessive diffusion. Besides, there remain missed instances. Further, an extreme case the system would confront is that a \textbf{Nov}el category that \textbf{D}iffers from the families of all known categories, termed \textbf{Nov-D}, may appear in the scene. This kind of data looks very different and is challenging. For example, the 'penguin' standing at the mountain is falsely classified as 'mountain bike' and, the corresponding CAM attends to multiple image regions and even backgrounds.


\begin{figure}[t]
	\centering
    \includegraphics[width=\linewidth]{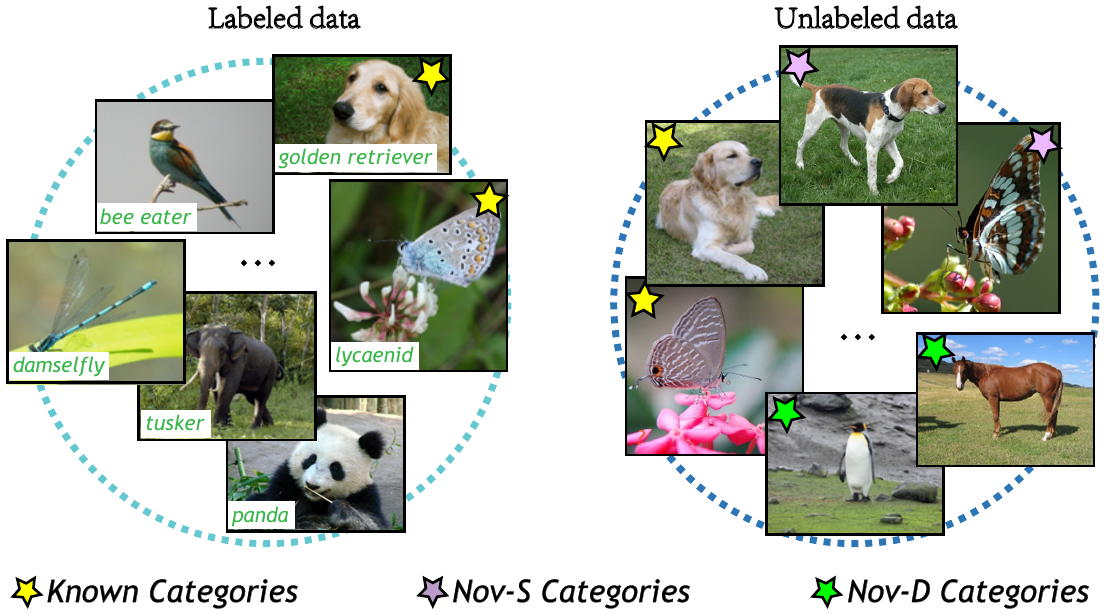}
	\vspace{-15pt}
	\caption{Illustration of the data partition of OWSOL. } 
	\label{fig:fig2}
 \vspace{-20pt}
\end{figure}


To mitigate the problem, we introduce a new weak localization task called Open-World Weakly-Supervised Object Localization (\textbf{OWSOL}). Beyond mitigating manual annotation efforts, OWSOL aims to motivate effective discovery and localization of both known and novel categories.



\textit{\textbf{Definition}}. Given a collection of training data comprising of labeled and unlabeled images, the task aims to learn generalized patterns for classifying and localizing known and novel objects in supervised and self-supervised manners. Here, all the labeled images in the training set come from \textbf{known categories}. Both the known and \textbf{novel ones} will appear in the unlabeled set. As aforementioned, some of novel objects may not share similar appearances to the known categories, which are challenging for WSOL. Thus, we further group novel categories into two parts according to the family overlap with known classes. As shown in Figure~\ref{fig:fig2}, if the family of a novel object, e.g., `English foxhound', is the same as one of the known classes, e.g., `golden retriever' (they belong to \textit{Canidae}), the object will be considered as a \textbf{novel} category in the \textbf{same} family to the known, i.e., \textbf{Nov-S}. If there is no family intersection between the novel and known, the object will be assigned to a \textbf{novel} category in a \textbf{different} family to the known, i.e., \textbf{Nov-D}. Note that such a partition is only available in \textit{test} data for evaluation. As a general setting to the real world, not all categories can be accessible and labeled to build an open-world object localization system, and there remain amounts of unlabeled data from both known and novel categories to be explored. Besides, the objects in the images may come from novel ones with significantly different appearances to the known ones (whether they belong to the same family or not), which may considerably degrade system performance. 






\textit{\textbf{Paradigm}}. Commonly, the parametric classification head used in closed-set WSOL is restricted to a fixed number of known categories and the learned feature space is susceptible to performance deterioration on novel data. Besides, such a parametric system is infeasible to unlabeled data. In this paper, we propose a simple yet effective co-learning paradigm to cope with both labeled and unlabeled data. In particular, given a mini-batch of labeled and unlabeled images, supervised contrastive learning is directly applied to the extracted representations.  Meanwhile, to prevent the features from overfitting to the subset of labeled categories, we perform the proposed semantic centroids-driven contrastive learning on both labeled and unlabeled images, which expands a well-structured feature space for known and novel categories. Specifically, clustering is conducted on the representations of all training samples to find semantic centroids and constitute a memory bank, i.e., each sample is assigned to a semantic centroid. In practice, the number of clusters is controlled to be larger than the number of underlying categories. Thus, various semantic centroids potentially focus on different salient object parts. Based on this assumption, multiple semantic centroids nearest to the learned representation are considered as the positive samples, and the other centroids are set as the negative samples. As representations are forced to be associated with various object regions, it would benefit the completeness of class activation map for object localization. 

In inference without a parametric classification head, predictions are accomplished by clustering on the learned representations. Because there are no class-specific weight vectors, it is implausible to get CAM, as in WSOL. Alternatively, we propose generalized activation mapping (G-CAM). Unlike the conventional CAM which computes activation map for each class (might be unknown in open world), we instead compute activation map for each cluster. This is achieved by sliding the semantic centroid vector over the feature map to compute their dot product similarity at each position of the feature map.


Comparisons in Figure~\ref{fig:fig1} (a) show that the proposed method can make accurate predictions and correctly attend to the complete object regions. Figure~\ref{fig:fig1} (b) underlines the significant improvements, across Known, Nov-S, and Nov-D categories, achieved by the proposed method.

The main contributions of this paper are summarized as:
\begin{itemize}
    \item We are the first to explore weakly-supervised object localization in open-world situations and introduce Open-World Weakly-Supervised Object Localization (OWSOL) task to motivate learning of weak localization models scalable to the open world.

    \item We propose a simple yet effective paradigm, including supervised and semantic centroids-driven contrastive co-learning, for OWSOL. Additionally, generalized class activation mapping (G-CAM) is proposed for object localization in a non-parametric manner. Extensive experiments show that the proposed paradigm surpasses the baseline by a large margin.
    
    \item We re-organize two widely used datasets, i.e., ImageNet-1K and iNatLoc500, and propose OpenImages150 to serve as evaluation benchmarks for OWSOL. The split of Nov-D categories is so far the most challenging part to validate the OWSOL methods.

\end{itemize}

\section{Related Works}

\textbf{Weakly-Supervised Object Localization}. Existing WSOL works usually employ image-level tags as weak supervision to learn a localization model. Zhou \etal~\cite{cam} proposed class activation mapping (CAM) for object localization, serving as the cornerstone in WSOL. However, the most distinct deficiency of CAM for localization is the concentrated activation of discriminative object regions, e.g., the head of a bird. To tackle this problem, various methods~\cite{adl, spg, acol, has, ewsol, RCAM, Pan_2021_CVPR, STL, Wei_2021_CVPR, FAM, TS-CAM, kim2021normalization, lctr, Kim_2022_CVPR, OLDA, BAS, CREAM, C2AM, } have been proposed. For example, Choe \etal~\cite{adl} devised an erasing technique to hide the most discriminative parts and stochastically highlight the informative regions, which can efficiently expand the attention of the model on objects.  However, the above weak localization methods are restricted to a small and fixed number of categories in a closed-world assumption. As mentioned in \cite{ossod}, out-of-distribution or novel objects would potentially fail to be classified and localized by the closed-set model. 


\begin{figure*}[ht]
	\centering
    \includegraphics[width=\linewidth]{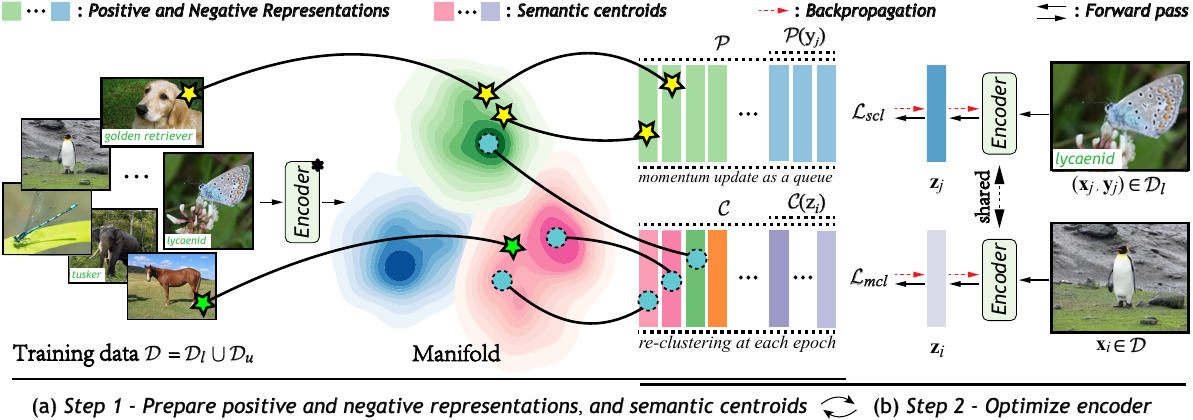}
	\vspace{-15pt}
	\caption{An overview of the proposed paradigm. (a) In the beginning, training data $\mathcal{D}$ will be mapped to representations by a pre-trained encoder like MoCo~\cite{moco}. Then, we construct a memory bank that stores a set of representations to provide sufficient positive and negative samples for the following representation learning. It is momentum updated as a queue at each iteration. Besides, clustering will be performed on the representation set to find semantic centroids. This constitutes a memory bank updated by re-clustering at the beginning of each epoch. * indicates that we use the momentum encoder to extract representations at each clustering (b) During training, given a mini-batch of labeled and unlabeled images, supervised contrastive learning $\mathcal{L}_{scl}$ and multiple centroids-driven contrastive learning $\mathcal{L}_{mcl}$ will be applied to learn representations. Best viewed in color.} 
	\label{fig:method}
    \vspace{-15pt}
\end{figure*}

\textbf{Open-World Visual Recognition}. In recent years, more and more works have been proposed to shift visual recognition towards an open-world setting. Han \etal~\cite{dtc} proposed a two-stage learning framework to transfer knowledge from known categories to unknown categories. Recent studies like NCL~\cite{ncl} and GCD~\cite{gcd} incorporated representation learning to well structure known and novel data. This work was motivated by the setting of generalized category discovery~\cite{gcd}, in which unlabeled training data involves both known and novel classes. Compared to the recognition of novel categories considered in the above methods, this work extends novel categories discovery to the localization problem and further partition novel categories as Nov-S and Nov-D for evaluation.  Beyond global recognition, there are many region- and pixel-level recognition, e.g., open-set detection~\cite{owod, ossod} and segmentation~\cite{OWS}. However, these works tend to tag novel categories as 'unknown' without any discovery or only segment the novel objects without classification. In comparison, this work tries to discover and localize novel categories at the same time.

\textbf{Contrastive Representation Learning}. Recent literature demonstrates the effectiveness of representation learning by contrasting positive pairs against negative pairs. Wu \etal~\cite{nid} proposed non-parametric instance discrimination by employing noise contrastive estimation and introducing a memory bank to store the instance class representations. Instead, Chen \etal~\cite{simclr} proposed to employ large batch sizes, and He \etal~\cite{moco} introduced a dynamic dictionary with a queue to produce sufficient negative samples. Beyond instance discrimination of different views, Li \etal~\cite{pcl} proposed EM-based prototypical learning to leverage multiple granularity information from clusters. Differently, Khosla \etal~\cite{supcon} employed the label information and proposed supervised contrastive learning by contrasting intra-class and inter-class representations. In contrast with the above methods, this work applies contrastive learning to both labeled and unlabeled data and takes into account the complete activation of objects in the loss function.


\section{Problem Setup}
Given a training set $\mathcal{D} = \{\vx_{i}\}^{n+m}_{i=1}$, where $\vx_{i} \in \mathcal{X}$, it consists of i) the labeled set $\mathcal{D}_l=\{\vx_{j},y_j\}^n_{j=1}$, where $y_j \in \mathcal{Y}_l$ and $\mathcal{Y}_l=\{\text{\underline{cat}}^1,\cdots, \text{\underline{cat}}^i, \cdots \}$ denotes the label set of \textbf{known} \underline{cat}egories; ii) unlabeled set $\mathcal{D}_u=\{\vx_{k}\}^{n+m}_{k=n+1}$, where each sample $\vx_k$ belongs to either known or \textbf{novel} categories. Novel categories in $\mathcal{D}_u$ may belong to the \textbf{Nov-S} or \textbf{Nov-D} categories. Note that, partition of Nov-S and Nov-D is only available in \textit{test} set for evaluation. For convenience, we denote the whole label set as $\mathcal{Y}_{all}$, where $\mathcal{Y}_l \in \mathcal{Y}_{all}$. Accordingly, the label set of novel categories $\mathcal{Y}_n = \mathcal{Y}_{all} \backslash \mathcal{Y}_{l}$. Further, $\mathcal{Y}_n = \mathcal{Y}_{s} \cup \mathcal{Y}_{d}$, where $\mathcal{Y}_{s}$ and $\mathcal{Y}_{d}$ imply the label set for the Nov-S and Nov-D categories. The data partition of OWSOL simulates a general scene where large amounts of unlabeled data may be available to be jointly employed with labeled data for training. It aims to motivate co-learning using labeled and unlabeled data to expand a more general feature space, in which known categories and novel ones are well structured.

After training, the learned model is applied to the whole \textit{test} data to extract a set of features. Clustering is then performed to associate each image to an underlying category in $\mathcal{Y}_{all}$ and class activation maps are generated to obtain bounding boxes for object localization.

\section{Methodology}
\label{sec:method}
In this section, we propose a simple yet effective paradigm to deal with OWSOL, which can serve as a foundation for subsequent works. In particular, we will introduce the proposed supervised and semantic centroids-driven contrastive co-learning in Section~\ref{sec:scl}, and the details of inference procedures including clustering and generalized class activation mapping (G-CAM) are presented in Section~\ref{sec:gcam}. In addition, we consider the estimation of number of underlying classes in Section~\ref{sec:estimation} and construct baselines for comparison from the literature in Section~\ref{sec:baseline}.

\subsection{Preliminary and System Framework}
\label{sec:pre}
We consider a base deep neural network encoder $f(\cdot) : 
\mathcal{X} \rightarrow \mathbb{R}^{d_1}$ that extracts representation vectors from data samples, i.e., $\vh = f(\vx)$. Besides, a MLP-based projection head $g(\cdot)$ in \cite{simclr} is used to map representations to space where contrastive learning is applied, i.e., $\vz=g(\vh)$ and $\vz \in \mathbb{R}^{d_2}$. Self-supervised pre-trained models like MoCo~\cite{moco} can be adopted to initialize the base encoder and projection head. For brevity, we use 'encoder' to represent the encoder with a MLP-based projection head. 

Before training, we first use the pre-trained encoder to extract representations $\{\vz_{i}\}^{n+m}_{i=1}$ of all samples in $\mathcal{D}$. For each known category $\text{\underline{cat}}^i$ in $\mathcal{Y}_{l}$, we randomly sample $N_z$ representations from $\{\vz_j\}^n_{j=1}$ to constitute a set  $\mathcal{P}(\text{\underline{cat}}^i) =\{\vz^1,\cdots, \vz^{N_z}\}$, which consequently forms a memory bank $\mathcal{P}=\{\mathcal{P}(\text{\underline{cat}}^1)\cup \cdots \cup \mathcal{P}(\text{\underline{cat}}^i)\cup \cdots \cup \mathcal{P}(\text{\underline{cat}}^{|\mathcal{Y}_l|})\}$. Given a sample with label $y_j$, we use $\mathcal{P}(y_j)$ to indicate the positive set of $N_z$ representations belonging to the same category with $y_j$. Meanwhile, clustering is performed on $\{\vz_{i}\}^{n+m}_{i=1}$ to obtain a large set of $N_c$ semantic centroids, constructing a memory bank $\mathcal{C}=\{\vc^1, \cdots, \vc^i, \cdots, \vc^{N_c}\}$. We use $\mathcal{C}({\vz_i})=\{\vc_{\vz_i}^1,\cdots, \vc_{\vz_i}^l, \cdots, \vc_{\vz_i}^L\}$ to represent $L$ semantic centroids associated with $\vz_i$, where $\vc_{\vz_i}^l$ is the $l$-th closest centroid to $\vz_i$. An illustration is depicted in Figure~\ref{fig:method} (a) and, the preparation and following training framework are presented in Algorithm~\ref{algo:a1}.


\subsection{Contrastive Representation Co-learning}
\label{sec:scl}
Given an anchor point $\vx_j$ with label $y_j$ in a mini-batch sampled from $\mathcal{D}_l$, the \textbf{s}upervised \textbf{c}ontrastive \textbf{l}oss is formulated as:
\begin{equation}
\label{eq:eq1}
    \mathcal{L}_{scl} = -\frac{1}{|\mathcal{P}(y_j)|} \sum_{\vz_j^+ \in \mathcal{P}(y_j)} \text{log} \frac{\text{exp}(\vz_j^{\top} \cdot \vz_j^+ / \tau)}{\sum_{\vz^i\in \mathcal{P}} \text{exp}(\vz_j^\top \cdot \vz^i / \tau)},
\end{equation}
where $\top$ is the matrix transpose and $\vz_j^+$ is the representation of positive sample to $\vz_j$. $\tau$ is the temperature parameter. Empirically, $\tau$ is set as 0.007. To keep representations consistent with the evolving encoder, $\mathcal{P}(y_j)$ is maintained as a queue in a momentum update at each iteration: the representations from the current mini-batch are enqueued into the corresponding positive set and the oldest are dequeued. This allows us to obtain sufficient positive and negative samples without requiring a large mini-batch size.

\begin{algorithm}[t]
\caption{Contrastive Representation Co-learning}
\label{algo:a1}
\begin{algorithmic}[1]
\State \textbf{Preparation:}
\State Extract representations $\{\vz_{i}\}^{n+m}_{i=1}$ of all samples in $\mathcal{D}$ using the pre-trained encoder.
\State  Initialize the memory bank of representations $\mathcal{P}$ and perform clustering on $\{\vz_{i}\}^{n+m}_{i=1}$ to construct the memory bank of semantic centroids $\mathcal{C}$.
\State \textbf{Main loop:}
\For{$t \in \{1, \dots, max\_epoch\}$} 
  \For{$i \in \{1, \dots, max\_iteration\}$}
    \State Train $f(\cdot)$ and $g(\cdot)$ using $\mathcal{L}_{scl}$ and $\mathcal{L}_{mcl}$.
    \State Momentum update $\mathcal{P}$ as a queue.
  \EndFor
  \State 
   Re-extract $\{\vz_{i}\}^{n+m}_{i=1}$ using the momentum encoder.
   \State 
   Re-clustering on $\{\vz_{i}\}^{n+m}_{i=1}$ to update $\mathcal{C}$. 
\EndFor
\end{algorithmic} 
\end{algorithm}

Given an anchor point $\vx_i$ in a mini-batch sampled from $\mathcal{D}$, the \textbf{o}ne semantic \textbf{c}entroid-driven contrastive \textbf{l}earning is conducted as:
\begin{equation}
\label{eq:eq2}
    \mathcal{L}_{ocl} = - \text{log} \frac{\text{exp}(\vz_i^{\top} \cdot \vc_{\vz_i}^1 / \phi(\vc_{\vz_i}^1))}{\sum_{\vc^i\in \mathcal{C}} \text{exp}(\vz_i^\top \cdot \vc^i / \phi(\vc^i))},
\end{equation}
where $\phi(\cdot)$ is a density estimation. Given the momentum features $\{\vz^v\}_{v=1}^V$ belonging to the same cluster with a semantic centroid $\vc^i$, we can estimate a density as follows:
\begin{equation}
    \phi(\vc^i) = \frac{\sum_{v=1}^V \left\|\vz^v-\vc^i\right\|_2}{V},
\end{equation}
where $V$ is the number of representations within the cluster. The smaller $\phi$ means a higher quality of the cluster because 1) the average distance between $\vz^v$ and $\vc^i$ is small and 2) the number of representations $V$ in the cluster is relatively large. Accordingly, we employ $\phi$ as the adaptive temperature to re-weight contrastive pairs with different quality clusters.

Equation~\ref{eq:eq2} means that each data point can only be pulled to approach a single semantic centroid. However, such a semantic centroid may have different emphases on different object regions due to the diverse appearances of objects. To learn representations with more granularity information, we employ multiple nearest semantic centroids as the positive templates for each anchor point. Accordingly, the \textbf{m}ultiple semantic \textbf{c}entroids-driven contrastive \textbf{l}oss can be formulated as follows: 


\begin{equation}
    \mathcal{L}_{mcl} = - \text{log} \frac{\text{exp}(\vz_i^{\top} \cdot \vc_{\vz_i}^* )}{\text{exp}(\vz_i^{\top} \cdot \vc_{\vz_i}^* ) + \sum\limits_{\vc^i\in {\mathcal{C} \backslash} \mathcal{C}({\vz_i})} \text{exp}(\vz_i^\top \cdot \vc^i / \phi(\vc^i))},
\end{equation}

\begin{equation}
    \vc_{\vz_i}^* = \frac{1}{L}\sum_{l=1}^{L}\vc_{\vz_i}^l/ \phi(\vc_{\vz_i}^l),
\end{equation}
where $\vc_{\vz_i}^l$ denotes the $l$-th closest semantic centroid to $\vz_i$ and $L$ is a hyper-parameter that controls the number of positive centroids to be approached.

The overall contrastive representation co-learning loss is the weighted summation of $\mathcal{L}_{scl}$ and $\mathcal{L}_{mcl}$:
\begin{equation}
\label{eq:eq6}
    \mathcal{L} = \alpha\mathcal{L}_{scl} + \beta\mathcal{L}_{mcl},
\end{equation}
where $\alpha$ and $\beta$ are hyper-parameters of two loss terms.


\subsection{Generalized Class Activation Mapping}
\label{sec:gcam}
Once the encoder is trained, we denote $\vm \in \mathbb{R}^{d_1\times h\times w}$ as the feature map extracted from the encoder $f(\cdot)$ without spatial pooling. Given a pre-trained closed-set classification model, there is a learned matrix $\vw \in \mathbb{R}^{d_1\times k}$ in the parametric classification layer. $h, w$, and $k$ are the spatial dimensions of the feature map and the number of pre-defined categories, respectively. Then, CAM (class activation map) for the $k$-th category can be derived as follows:
\begin{equation}
    \vp_k(i,j) = \vw_k^{\top}\vm(i,j),
\end{equation}
where $\vw_k \in \mathbb{R}^{d_1\times 1}$ is the $k$-th learned vector of $\vw$, and $\vm(i,j) \in \mathbb{R}^{d_1\times 1}$ indicates the representation vector located on $(i,j)$. Though it is an effective manner to give spatial activation to those known categories, they are not applicable to novel ones without corresponding learned vectors. Thus, we propose the generalized class activation mapping (G-CAM), in which clustering assignments and semantic centroids are employed to replace the learnable matrix with fixed size. Given a collection of test data, we first extract the representation vectors using the trained encoder and estimate the number of underlying classes. Clustering is then performed on these representations to associate each image with a semantic centroid. Given an image $\vx$, assuming $\vc_\vx \in \mathbb{R}^{d_1 \times 1}$ is the semantic centroid assigned to $\vx$ and the G-CAM can be obtained as follows:
\begin{equation}
    \vp_\vx(i,j) = \vc_\vx^{\top}\vm(i,j).
\end{equation}
It is a general version of CAM that shifts closed-set activation mapping towards the open-world problem. After getting G-CAM, a thresholding procedure is used for binarization, and the object bounding box can be extracted from the largest connected component.

\begin{table}[b]
	\centering
     \vspace{-5pt}
	\caption{Datasets summarization.}
	\resizebox{\linewidth}{!}{
		\begin{tabular}{l c c c c c c c }
		\toprule[1.5pt]
        \rowcolor{mygray}
		 & $|\mathcal{D}_l|$ & $|\mathcal{D}_u|$ & $|\mathcal{Y}_l|$ & $|\mathcal{Y}_s|$ & $|\mathcal{Y}_d|$ & \textit{$|$val$|$} & \textit{$|$test$|$} \\
		\midrule
             ImageNet-1K & 320,122 & 961,045 & 500 & 250 & 250 & 10,000 & 50,000 \\
             iNatLoc500 &  34,589 & 103,405 & 250 & 125 & 125 & 12,500 & 12,500  \\
             OpenImages150 &  11098 & 33370 & 75 & \multicolumn{2}{c} {75} & 7,500 & 3,750 \\
		\bottomrule[1pt]
	\end{tabular}}
	\label{tab:dataset_summ}
	\vspace{-10pt}
\end{table}

\begin{table*}[ht] \footnotesize
	\centering
	\caption{Evaluation results on ImageNet-1K. The best results are in \textbf{bold}.}
	\resizebox{\linewidth}{!}{
		\begin{tabular}{l c c c c c c c c c c c c c c }
		\toprule[1pt]
		\multirow{2.5}{*}{Methods} &\multicolumn{4}{c}{Clus Acc} &\multicolumn{4}{c}{Clus-Loc Acc} & \multicolumn{4}{c}{Loc Acc }\\
		\cmidrule(lr){2-5} \cmidrule(lr){6-9} \cmidrule(lr){10-13}
		 & Known & Nov-S & Nov-D & All & Known & Nov-S & Nov-D & All & Known & Nov-S & Nov-D & All \\
		\midrule
            CAM~\cite{cam}            & 33.55  & 19.05 & 17.84 & 26.00 & 22.94 & 12.84 & 11.25 & 17.49 & 58.04  & 59.25 & 53.18 & 57.13  \\
            HaS~\cite{has}            & 32.86  & 19.62 & 18.41 & 25.94 & 22.55 & 13.46 & 11.79 & 17.59 & 58.33  & 59.72 & 53.48 & 57.46  \\
            SPG~\cite{spg}            & 33.52  & 18.33 & 17.40 & 25.69 & 22.87 & 12.10 & 10.81 & 17.16 & 58.05  & 59.29 & 53.88 & 57.32  \\
            ADL~\cite{adl}            & 33.30  & 19.02 & 17.25 & 25.72 & 22.88 & 12.71 & 10.89 & 17.34 & 58.73  & 59.66 & 53.86 & 57.74  \\
            WSOL-DA~\cite{OLDA}            & 37.72 & 19.06 & 17.52 & 28.00 & 25.70 & 13.31 & 11.07 & 18.95 & 58.76 & 59.99 & 53.39 & 57.72  \\
            BAS~\cite{BAS}            & 47.33 & 28.11 & 27.34 & 37.53 & 31.84 & 19.67 & 17.65 & 25.25 & 60.53 & 63.51 & 55.26 & 59.96  \\
            \midrule
            GCD~\cite{gcd}             & 51.57 & 22.62 & 21.12 & 36.72 & 35.65 & 15.09 & 12.88 & 24.82 & 61.68 & 62.07 & 54.28 & 59.93 \\

		
		\midrule
        \rowcolor{mygray}
        \textbf{G-CAM (Ours)}           & \textbf{58.64} & \textbf{36.77} & \textbf{34.78} & \textbf{47.20} &\textbf{39.21} &\textbf{25.79} &\textbf{22.82} &\textbf{31.76} &\textbf{61.75} &\textbf{65.11} &\textbf{57.60} &\textbf{61.56}   \\
		
		\bottomrule[0.5pt]
	\end{tabular}}
	\label{tab:imagenet1k}
	\vspace{-10pt}
\end{table*}

\begin{table*}[ht] \footnotesize
	\centering
	\caption{Evaluation results on iNatLoc500. The best results are in \textbf{bold}.}
	\resizebox{\linewidth}{!}{
		\begin{tabular}{l c c c c c c c c c c c c c c }
		\toprule[1pt]
		\multirow{2.5}{*}{Methods} &\multicolumn{4}{c}{Clus Acc} &\multicolumn{4}{c}{Clus-Loc Acc} & \multicolumn{4}{c}{Loc Acc }\\
		\cmidrule(lr){2-5} \cmidrule(lr){6-9} \cmidrule(lr){10-13}
		 & Known & Nov-S & Nov-D & All & Known & Nov-S & Nov-D & All & Known & Nov-S & Nov-D & All \\
		\midrule
            CAM~\cite{cam}            & 27.33  & 18.43 & 15.01 & 22.02  & 15.79 &9.92 & 7.26 & 12.19 & 45.94  & 44.66 & 41.78 & 44.58  \\
            HaS~\cite{has}            & 25.98  & 17.09 & 14.75 & 20.95 & 14.62 & 9.06 & 6.69 & 11.25 & 44.55  & 43.54 & 39.59 & 43.06  \\
            SPG~\cite{spg}            & 27.86  & 18.75 & 14.46 & 22.23 & 14.91 & 9.92 & 6.40 & 11.54 & 42.97  & 43.65 & 40.55 & 42.54  \\   
            ADL~\cite{adl}            & 24.94  & 18.82 & 14.91 & 20.90 & 13.88 & 10.13 & 6.71 & 11.15 & 45.05 & 44.77 & 41.33 & 44.05  \\
            WSOL-DA~\cite{OLDA}            & 27.94 & 19.07 & 15.04 & 22.50 & 15.57 & 9.73 & 6.27 & 11.78 & 45.07 & 44.74 & 38.67 & 43.38  \\
            BAS~\cite{BAS}            & 35.36 & 26.24 & 20.32 & 29.32 & 23.19 & 17.32 & 11.68 & 18.85 & 56.98 & 57.21 & 54.56 & 56.43  \\
            \midrule
            DTC~\cite{dtc}                      & 24.91 & 19.55 & 15.17 & 21.14 & 15.79  & 12.12 & 7.58 & 12.82  & 52.36 & 53.95 & 46.89 & 51.39\\
            GCD~\cite{gcd}                     & 36.74 & 23.10 & 17.70 & 28.57 & 26.77  & 15.13 & 9.46 & 19.53  & 61.41 & 57.67 & 46.79 & 56.82  \\ 
		
		\midrule
        \rowcolor{mygray}
		\textbf{G-CAM (Ours)}           & \textbf{47.76} & \textbf{29.73} & \textbf{21.95} & \textbf{36.80} &\textbf{33.36} &\textbf{21.34} &\textbf{14.61} &\textbf{25.67} &\textbf{63.67} &\textbf{63.29} &\textbf{60.86} &\textbf{62.87}   \\
		
		\bottomrule[0.5pt]
	\end{tabular}}
	\label{tab:inatloc500}
	\vspace{-10pt}
\end{table*}

\subsection{Baseline}
\label{sec:baseline}
To the best of our knowledge, this paper is the first work to handle the open-world problem of weakly supervised object localization. Thus, there is no baseline for the proposed method to compare from the literature. As an alternative, we directly adapt some classical works of WSOL, i.e., CAM~\cite{cam}, HaS~\cite{has}, SPG~\cite{spg}, ADL~\cite{adl}, DA-WSOL~\cite{OLDA}, and BAS~\cite{BAS} to the proposed OWSOL setting. Note that these methods cannot employ the available unlabeled data because of the parametric classification head. For a fair comparison, the predictions are obtained by clustering, and the generalized class activation mapping is used for object localization. Besides, novel categories discovery works like DTC~\cite{dtc} and GCD~\cite{gcd} are adapted as the baseline.

\subsection{Number of Underlying Classes}
\label{sec:estimation}
For convenience, we assume that the number of underlying classes in the unlabeled data $\mathcal{D}_u$ is known to compute accuracy across our experiments. However, it is unrealistic that the prior knowledge about the number of classes is available. Therefore,  following~\cite{dtc, gcd}, we attempt to estimate the number of underlying classes in the unlabeled data and probe the variations of performance with such an operation. The experimental results are included in Table~\ref{tab:estimation}. 

\section{Experiments}
\label{sec:experiment}
\subsection{Experimental setup}
\textbf{Datasets}. We re-organize two commonly used datasets, i.e., ImageNet-1K~\cite{imagenet1k} and iNatLoc500~\cite{inatloc500}, and propose OpenImages150 to serve as evaluation benchmarks for OWSOL. A summarization is presented in Table~\ref{tab:dataset_summ}. For ImageNet-1K, we manually select 500 categories as the known and the remaining 500 categories as the novel ones. Besides, the novel categories are further split into 250 Nov-S and 250 Nov-D categories. There are 320,122 and 961,045 images in the labeled and unlabeled set, respectively, for training. 10,000 and 50,000 images are available for validation and test, respectively. For iNatLoc500~\cite{inatloc500}, 250 categories are manually selected as the known and the remaining is split into 125 Nov-S and 125 Nov-D categories. There are 34,589 and 103,405 images in the labeled and unlabeled set, respectively, for training. 10,000 and 50,000 images are available for validation and test, respectively. We also propose a new dataset, i.e., OpenImages150 for OWSOL. We sub-sample 150 categories with bounding box annotations from the original OpenImages dataset\cite{openimages}. 75 categories are randomly adopted as the known and the remaining categories are used as the novel. More details of these datasets are shown in the supplementary materials.

\textbf{Evaluation Metrics}. Motivated by previous studies~\cite{imagenet1k, cam, ewsol}, we introduce three metrics, clustering accuracy (Clus Acc), clustering and localization accuracy (Clus-Loc Acc), and localization accuracy (Loc Acc), for evaluation. Clustering accuracy is evaluated by running the Hungarian algorithm~\cite{hungarian} to find the optimal assignment between the set of cluster indices and groud-truth labels. A localization is correct when the predicted bounding box overlaps over a ratio with one of the ground truth bounding boxes belonging to the same class. For Loc Acc, we average the localization performance across different bounding box ratios, i.e., \{30\%, 50\%, 70\%\}. For Clus-Loc Acc, a prediction is correct when the clustering and localization results are both correct and it is also averaged across three different bounding box overlap ratios, i.e., \{30\%, 50\%, 70\%\}.

\begin{table}[t]
	\centering
 \vspace{-5pt}
	\caption{Evaluation results on OpenImages150.}
	\resizebox{\linewidth}{!}{
		\begin{tabular}{l c c c c c c c c c}
		\toprule[1.5pt]
		\multirow{2.5}{*}{Methods} &\multicolumn{3}{c}{Clus Acc} &\multicolumn{3}{c}{Clus-Loc Acc} & \multicolumn{3}{c}{Loc Acc }\\
		\cmidrule(lr){2-4} \cmidrule(lr){5-7} \cmidrule(lr){8-10}
		 & Known & Novel & All & Known & Novel & All & Known & Novel & All \\
		\midrule
            CAM~\cite{cam}            & 53.01 /  & 21.60  & 37.30  & 33.48  & 11.68  & 22.58  & 53.70 /  & 47.38  & 50.54   \\
            HaS~\cite{has}            & 51.06 /  & 21.09  & 36.08  & 31.71  & 11.46  & 21.58  & 53.79 /  & 46.79  & 50.28 \\
            SPG~\cite{spg}            & 56.45 /  & 24.93  & 40.69  & 35.58  & 12.57  & 24.07  & 53.63 /  & 46.65  & 50.14  \\
            ADL~\cite{adl}          & 50.05   & 22.13  & 36.09  & 30.77  & 11.52  & 21.15  & 52.74   & 46.40  & 49.57  \\
            WSOL-DA~\cite{OLDA}          & 53.73  & 26.27  & 40.00  & 32.05  & 13.66  & 22.86  &  52.59  & 45.77  & 49.18  \\
            BAS~\cite{BAS}          & 53.15  & 25.44  & 39.29  & 32.27  & 12.48  & 22.37  & 52.35   & 44.58  & 48.44  \\



            \midrule
            DTC~\cite{dtc} & 42.40 & 23.60  & 33.00  & 25.36  & 12.27  & 19.31  & 52.92  & 46.65  & 49.79  \\
            GCD~\cite{gcd} & 61.97 & 38.29  & 50.13  & 39.45  & 22.60  & 31.02  & 56.57  & 49.08  & 52.82 \\
		
		\midrule
        \rowcolor{mygray}
		\textbf{G-CAM (Ours)}           & \textbf{64.35} & \textbf{47.25} & \textbf{55.80} & \textbf{40.59} &\textbf{27.36} &\textbf{33.97}&\textbf{57.07}&\textbf{50.17}&\textbf{53.62}   \\ 
		
		\bottomrule[1pt]
	\end{tabular}}
	\label{tab:openimages150}
	\vspace{-20pt}
\end{table}

\textbf{Implementation Details}. For our method, we adopt SGD optimizer with a weight decay of 0.0001 and a momentum of 0.9. For supervised contrastive learning, $N_z$ is set as 12. For semantic centroids-driven contrastive learning, we set the number of centroids as 50,000 and 5,000 for ImageNet-1K and iNatLoc500, respectively. The hyper-parameters $\alpha$ and $\beta$ in Equation~\ref{eq:eq6} are set to 1.0 and 0.5, respectively. For ImageNet-1K dataset, we use a mini-batch size of 256 and an initial learning rate of 0.003.  For iNatLoc500 dataset, we use a mini-batch size of 128 and an initial learning rate of 0.001. More details can be found in supplementary materials.


\begin{table}[t]
	\centering
	\caption{Ablation studies on different loss combinations.}
	\resizebox{\linewidth}{!}{
		\begin{tabular}{c c c c c c c c c c}
			\toprule[1.5pt]
			\multirow{2.5}{*}{$\mathcal{L}_{scl}$} & \multirow{2.5}{*}{$\mathcal{L}_{ocl}$} & \multirow{2.5}{*}{$\mathcal{L}_{mcl}$} & \multirow{2.5}{*}{$\mathcal{L}_{ce}$} & \multicolumn{3}{c}{Clus Acc} & \multicolumn{3}{c}{Loc Acc } \\
            \cmidrule(lr){5-7} \cmidrule(lr){8-10}
            & & & &  Known & Nov-S & Nov-D  & Known & Nov-S & Nov-D \\ 
            \midrule
			\multicolumn{4}{l}{Zero-shot using MoCo~\cite{moco}} & 22.38 & 18.72 & 22.44 & 52.98 & 54.02 & 51.33 \\
            \midrule
			& & & \checkmark & 33.55 & 19.05 & 17.84 & 58.04 & 59.25 & 53.18 \\
			\midrule
			\checkmark  &   &   & & 51.02 & 20.41 & 17.43 & 58.55 & 60.46 & 53.84  \\
			\checkmark  & \checkmark  &   &  & \textbf{59.41}  & 35.54  & 33.19  & 58.95 & 61.75 & 55.29 \\
			\rowcolor{mygray}
			\checkmark  &    & \checkmark  &  & 58.64 & \textbf{36.77} & \textbf{34.78} &\textbf{61.75} &\textbf{65.11} &\textbf{57.60}   \\		
			
			\bottomrule[1pt] 
	\end{tabular}}
	\label{tab:ablation}
	\vspace{-15pt}
\end{table}

\subsection{Quantitative Results}
 Table~\ref{tab:imagenet1k} compares our method with the constructed baselines using classical WSOL works, including CAM~\cite{cam}, HaS~\cite{has}, SPG~\cite{spg}, and ADL~\cite{adl}, on ImageNet-1K dataset. One can observe from the table that CAM achieves 33.55\% Clus Acc and 22.94\% Clus-Loc Acc on known categories, respectively. As expected, the Clus and Clus-Loc Acc of CAM on novel categories are relatively lower than that of known categories. Compared with CAM, other baselines like HaS, SPG, and ADL mostly have degradation in Clus and Clus-Loc Acc on known and novel categories. For Loc Acc metric, CAM obtains 58.04\% on known categories. An interesting phenomenon we found is that the learned representations of CAM obtain a relatively higher Loc Acc, 59.25\%, on the Nov-S categories compared to the known ones. The reason is that the family of multiple categories in Nov-S is the same as one of the known categories. For example, there are 76 and 42 types of dogs in known and Nov-S categories, respectively. Thus, a large number of training samples within the same family in the known categories benefits the localization of dogs in Nov-S categories. Besides, many families, e.g., 'koala' and 'llama', only have one or two categories in the known group. The lack of this kind of data leads to difficulty in localizing these known categories. Thus, the Loc Acc on known categories is accordingly lower than that of Nov-S ones. More details can be found in supplementary materials. Other baselines like HaS, SPG, and ADL gain minor improvements of Loc Acc over CAM across different categories. 

 Compared with baseline from WSOL, GCD presents better performance across known, Nov-S, and Nov-D categories. Compared with GCD, our method can also immensely surpass it with a large margin across different metrics on different categories. Specifically, the Clus and Clus-Loc Acc of 'All' are improved from 36.72\% to 47.20\% and 24.82\% to 31.76\%, respectively. In addition, the proposed method achieves 61.75\%, 65.11\%, and 57.60\% Loc Acc on known, Nov-S, Nov-D categories, with improvements of 3.71\%, 5.86\%, 4.42\% over CAM, respectively. Comparisons on iNatLoc500 and 
OpenImages150 are presented in Table~\ref{tab:inatloc500} and \ref{tab:openimages150}. Clearly, the proposed method also demonstrates large improvements over the baselines across different metrics on different categories. 

Results for known categories using the parametric classification head are also listed after the slash '/'. Obviously, the parametric head has a good performance on known categories. However, features learned in such a manner failed to be generalized to open-world objects, i.e., Nov-S and Nov-D. In contrast, our non-parametric method  maintains competitive performance on known categories and shows an impressive ability to discover and localize novel objects.


\begin{figure}[t]
	\centering
    \includegraphics[width=\linewidth]{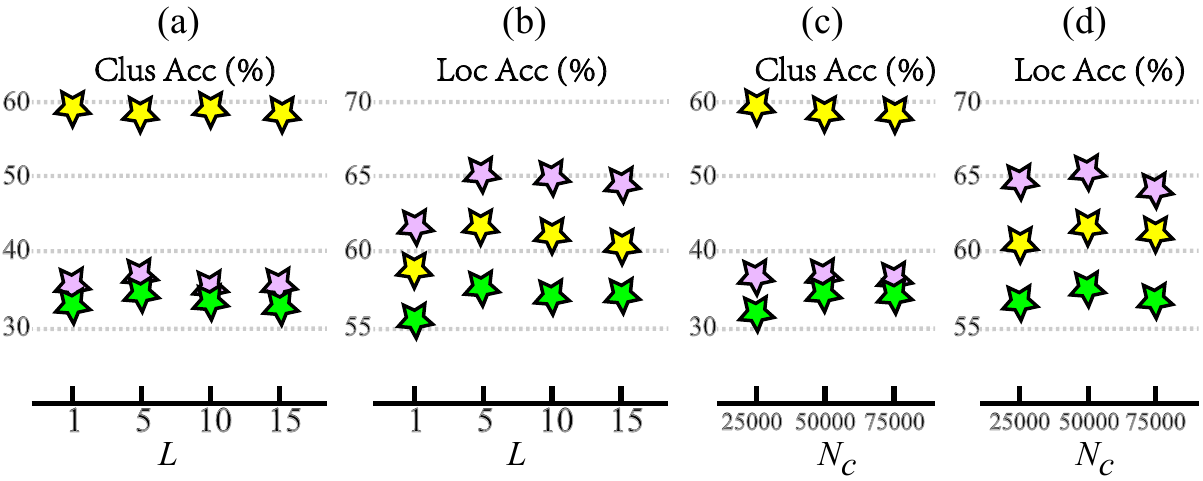}
	\vspace{-15pt}
    \caption{Analysis of the number of semantic centroids $L$ used in $\mathcal{L}_{mcl}$ and number of clusters $N_c$ used in each clustering. The results are for the ImageNet-1K. Stars in \textcolor{y}{yellow}, \textcolor{p}{purple}, and \textcolor{g}{green} indicate the result for Known, Nov-S, and Nov-D categories.} 
	\label{fig:ablation_centroids_clusters}
 \vspace{-10pt}
\end{figure}



\subsection{Ablation Studies}
The proposed method, as described in Section~\ref{fig:method}, is optimized by a combination of two loss functions, i.e., $\mathcal{L}_{scl}$ and $\mathcal{L}_{mcl}$. To examine the effectiveness of these two loss functions compared with $\mathcal{L}_{ocl}$ and the cross-entropy loss $\mathcal{L}_{ce}$ with a parametric classification head, we conduct ablation studies on these loss terms, and the experimental results on ImageNet-1K are presented in Table~\ref{tab:ablation}. When the encoder is followed by a parametric classification head and only supervised by $\mathcal{L}_{ce}$, the learned representations achieve only 33.55\% Clus Acc and 58.04\% Loc Acc for known categories. As expected, significant drops in Clus Acc can be found in novel categories, i.e., only 19.05\% Clus Acc on the Nov-S and a much lower Clus Acc of 17.84\% on Nov-D categories are achieved. This validates that the categories Nov-D are more challenging.  With only $\mathcal{L}_{ce}$, the learned representations only achieve 53.18\% Loc Acc on Nov-D categories, with a margin of about 6.0\% lower than that of the known and Nov-S ones. It suggests that the Nov-D objects are the most challenging to be localized. Alternatively, we remove the parametric classification head and adopt the proposed supervised contrastive learning $\mathcal{L}_{scl}$. One can observe that such an operation remarkably improves the Clus Acc on known categories while keeping close results to $\mathcal{L}_{ce}$ on novel categories, but only marginally raises the Loc Acc performance. When incorporating the unlabeled data for training using $\mathcal{L}_{ocl}$, considerable improvements of Clus Acc and Loc Acc are gained across known and novel categories. However, as the number of clusters is controlled to be far larger than the number of underlying categories, a single semantic centroid tends to focus on specific object regions. This subsequently affects the complete activation of objects in the activation maps. To this end, we devise $\mathcal{L}_{mcl}$ to adopt multiple nearest semantic centroids as the positive templates. Surprisingly, the involvement of $\mathcal{L}_{mcl}$ further improves Loc Acc across known, Nov-S, and Nov-D categories by 2.80\%, 3.36\%, and 2.31\%, respectively. 


\begin{figure}[t]
    \centering
    \includegraphics[width=\linewidth]{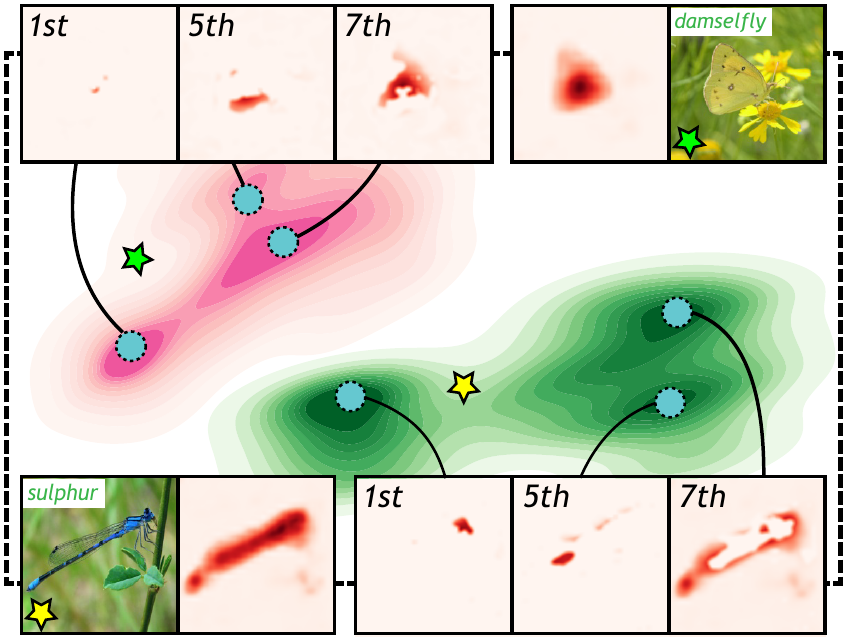}
    \caption{Probe of semantic centroids. Multiple semantic centroids nearest to the image are used to generate activation maps.} 
	\label{fig:probe}
 \vspace{-15pt}
\end{figure}


\textbf{Number of Centroids $L$ in $\mathcal{L}_{mcl}$}. We present the Clus and Loc Acc obtained using different numbers of semantic centroids in Figure~\ref{fig:ablation_centroids_clusters} (a) and (b). When the number of semantic centroids $L$ is set as 1, $\mathcal{L}_{mcl}$ equals $\mathcal{L}_{ocl}$ and it achieves the lowest Loc Acc. By contrast, the involvement of 5 semantic centroids in $\mathcal{L}_{mcl}$ considerably improves the localization potential of the learned representations on known and novel categories. Besides, such an involvement of multiple centroids in representation learning does not affect the Clus Acc. However, there is a performance deterioration when the number of semantic centroids is increased to 10 and 15. 

\textbf{Number of Clusters $N_c$}. Sensitivity analysis of the number of clusters $N_c$ used in each clustering are shown in Figure~\ref{fig:ablation_centroids_clusters} (c) and (d). One can observe that relatively stable performance of Clus Acc is obtained using different numbers of clusters. On the contrary, the learned representations using 50,000 clusters achieve the highest Loc Acc, which is better than that of 25,000 and 75,000 clusters. 

\subsection{Discussions}



\textbf{Probe of Semantic Centroids}. To investigate the inherent cause of improvements using multiple centroids in $\mathcal{L}_{mcl}$ and justify our assumption, we visualize the activation maps using various semantic centroids in Figure~\ref{fig:probe}. The samples are from known and Nov-S categories on ImageNet-1K. It can be seen that, for the known category 'sulphur', the closest ($1$-st) centroid to the sample is highly related to the body of the 'sulphur'. By contrast, the $5$-th and $7$-th closest semantic centroids focus on the long tail and head regions, respectively. Similar phenomena are also found in activation maps of Nov-S category 'damselfly'. This validates that different centroids focus on different object semantics. Thus, multiple semantic centroids involved in the representation learning will benefit the completeness of activation maps. 

\textbf{Number of Underlying Classes}. Table~\ref{tab:estimation} presents the performance of the proposed method when the number of underlying classes is estimated. It is clear that the estimated number of classes (884 and 519) is close to the ground-truth (1000 and 500). Surprisingly, the Clus Acc and Loc Acc of our method are similar to the results using the ground-truth number of classes in Table~\ref{tab:inatloc500}. 
\begin{table}[t]
	\centering
	\caption{Results when the number of classes is estimated. E / G indicates the estimation and ground-truth number of classes.}
	\resizebox{\linewidth}{!}{
		\begin{tabular}{c c c c c c c c}
			\toprule[1.5pt]
			 \multirow{2.5}{*}{Datasets} & \# Classes & \multicolumn{3}{c}{Clus Acc} & \multicolumn{3}{c}{Loc Acc } \\
             \cmidrule(lr){2-2} \cmidrule(lr){3-5} \cmidrule(lr){6-8}
            & E / G &  Known & Nov-S & Nov-D & Known & Nov-S & Nov-D \\ 
			\midrule
            ImageNet-1K & 884 / 1000 & 59.15 & 34.18 & 33.81  & 61.72 & 65.00 & 57.38 \\
             iNatLoc500 & 519 / 500 & 48.19 & 30.43 & 21.89  & 63.83 & 63.46 & 60.85 \\  
             
			\bottomrule[1pt] 
	\end{tabular}}
	\label{tab:estimation}
	\vspace{-15pt}
\end{table}

\textbf{Zero-shot Ability}. Table~\ref{tab:zero-shot} shows zero-shot ability to novel classes not seen in training for iNatLoc500. 250 novel classes are divided into 100 and 150 novel classes called Nov-100 and Nov-150, respectively. Only Known classes and Nov-100 are used for training and the remaining Nov-150 is adopted for zero-shot evaluation. Table~\ref{tab:zero-shot} shows that our G-CAM performs much better than CAM. The Clus Acc on Nov-150 is about 5\% lower when it's not seen in the training, while the Loc Acc is only slightly decreased from 60.88\% to 60.54\%. The results suggest that our method has a strong zero-shot ability to localize novel classes never seen before. More experimental results are presented in supplementary materials.

\begin{table}[h]
\vspace{-8pt}
 \caption{Zero-shot ability to classes not seen in training.}
 \resizebox{\linewidth}{!}{
  \begin{tabular}{c c c >{\columncolor{mygray}} c c >{\columncolor{mygray}}c}
   \toprule
   \multirow{2.5}{*}{Methods} & {seen in training (\checkmark) or not ($\times$)}
   & \multicolumn{2}{c}{Clus Acc} & \multicolumn{2}{c}{Loc Acc } \\
            \cmidrule(lr){2-2} \cmidrule(lr){3-4} \cmidrule(lr){5-6}
             & Nov-150  &Nov-100 & Nov-150 &Nov-100 & Nov-150 \\ 
            \midrule
            CAM  &$\times$ & 18.40 & 15.60 & 44.19 & 42.13  \\
            \rowcolor{mygray1}
            G-CAM  &$\times$ & 25.36 & 20.77 & 63.48 & 60.54  \\        
            G-CAM  &\checkmark & 26.40 & 25.47 & 62.76 & 60.88  \\ 
   \bottomrule
 \end{tabular}}
 \label{tab:zero-shot}
 \vspace{-18pt}
\end{table}

\section{Conclusions}
This paper proposes an open-world weak localization benchmark called Open-World Weakly-Supervised Object Localization (OWSOL) and re-organizes three commonly used datasets for evaluation. It motivates the learning of more scalable weak localization models towards open-world scenarios. The proposed simple yet effective paradigm with generalized class activation mapping demonstrates a better performance than baseline methods, which can serve as a strong foundation for subsequent works.

{\textbf{Limitation}. Partition of Nov-S and Nov-D can serve as an evaluation of novel categories with various difficulties. However, the proposed method treats them as unlabeled data in representation learning without differences. Fine-grained learning of Nov-S and Nov-D may help improve the capacity of model and we leave it to future works.}

{\small
\bibliographystyle{ieee_fullname}
\bibliography{egbib}
}


\twocolumn[
\centering
\Large
\textbf{Open-World Weakly-Supervised Object Localization} \\
\vspace{0.5em}Appendices \\
\vspace{1.0em}
{%
\renewcommand\twocolumn[1][]{#1}%
\maketitle
\begin{center}
\centering
\end{center}
}]

\begin{figure*}[!t]
	\centering
    \includegraphics[width=\linewidth]{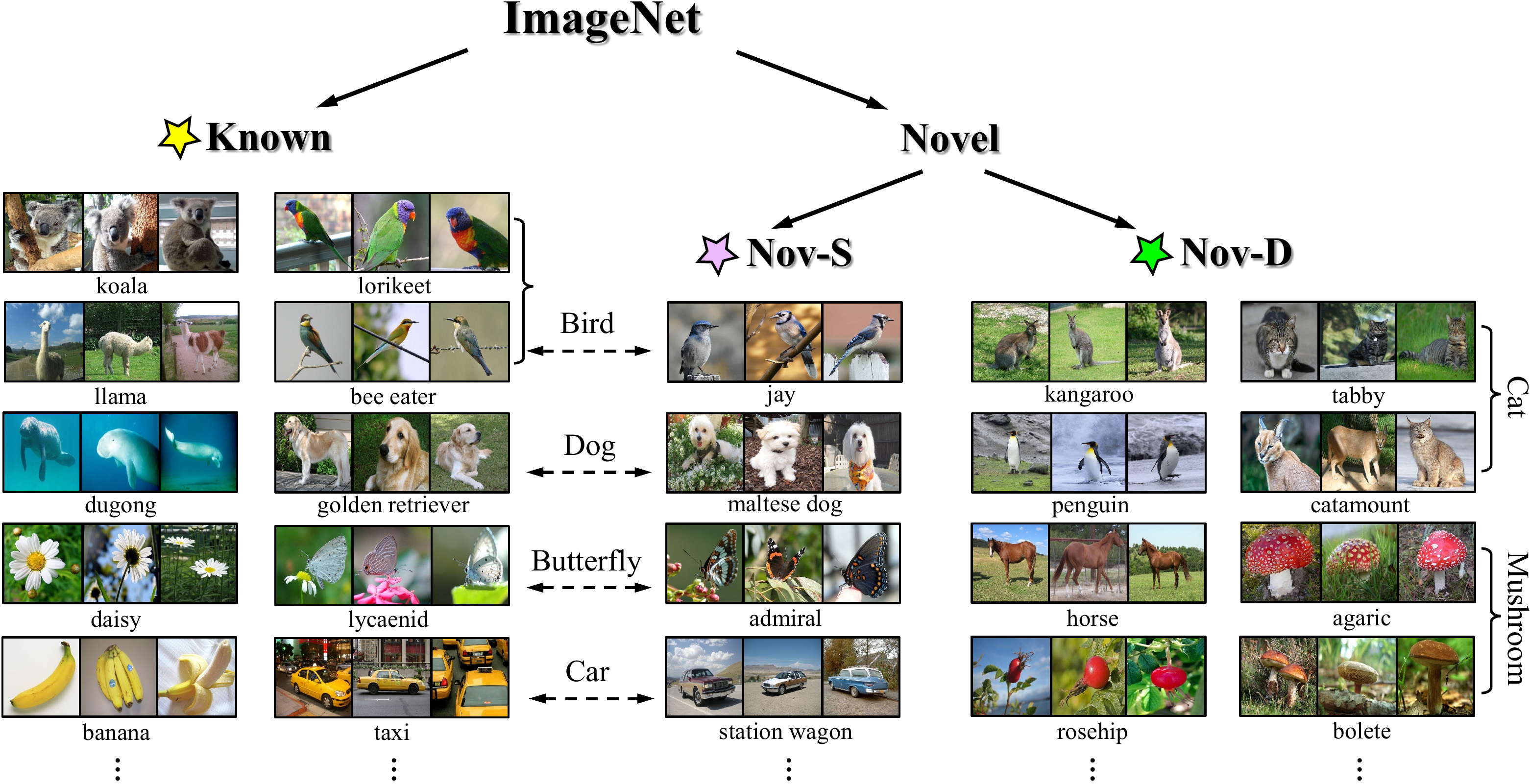}
    \vspace{-18pt}
	\caption{Samples of \textbf{Known}, \textbf{Nov-S}, and \textbf{Nov-D} categories on the re-organized ImageNet-1K.} 
	\label{fig:dataset_imagenet}
\end{figure*}


\appendix

\section{Data Samples of ImageNet-1K}
We present some examples of \textbf{Known}, \textbf{Nov-S}, and \textbf{Nov-D} categories in Figure~\ref{fig:dataset_imagenet}. In general, there is no overlap between known and novel categories. However, note that there are many fine-grained categories that belong to the same family like 'Bird' and 'Dog' between Known and Nov-S. These kinds of categories share resemblance in color, texture, or shape. Differently, objects in Nov-D look 
significantly different from Known and Nov-S categories, which raises the most challenging set for novel category discovery and localization. It is an analogy to the open world that the undiscovered data may contain categories with similar features or substantially different appearances  to the known ones. The re-organized ImageNet-1K, iNatLoc500, and the proposed OpenImages150 will be released.
\section{More Details about OpenIamges150}
We propose a new dataset, i.e., OpenImages150, to serve as an evaluation benchmark for OWSOL. In detail, we sub-sample 150 categories from the instance segmentation subsets of the official OpenImages dataset. The proposed OpenIamges150 includes 44,468, 7500, and 3750 images in the \textit{train}, \textit{test}, and \textit{val} set, respectively. Each category has about 300 training samples. To adapt OpenImages150 to the setting of OWSOL, 75 categories are partitioned as \textbf{Known} categories and the remaining 75 categories are treated as \textbf{Novel} categories. The partition and class names of OpenImages150 are presented below:

\noindent\fbox{%
	\parbox{0.97\linewidth}{%
		\textbf{Categories Partition of OpenImages150}
		
		\{\textcolor{blue}{'Known Categories'}: ['Apple', 'Boot', 'Shorts', 'Bus', 'Bicycle wheel', 'Dress', 'Vehicle registration plate', 'Limousine', 'Carnivore', 'Cattle', 'Cat', 'Bronze sculpture', 'Coin', 'Tart', 'Fox', 'Horse', 'Motorcycle', 'Lizard', 'Mushroom', 'Chest of drawers', 'Piano', 'Penguin', 'Pizza', 'Pig', 'Rabbit', 'Sculpture', 'Squirrel', 'Snake', 'Skyscraper', 'Tank', 'Tomato', 'Trousers', 'Lion', 'Bread', 'Frog', 'Shark', 'Flower', 'Hamburger', 'Fish', 'Airplane', 'Camera', 'Cake', 'Van', 'Pastry', 'Pen', 'Aircraft', 'Canoe', 'Chopsticks', 'Laptop', 'Clock', 'Camel', 'Mug', 'Tap', 'Vase', 'Giraffe', 'Jaguar', 'Sheep', 'Tiger', 'Strawberry', 'Whale', 'Duck', 'Otter', 'Bull', 'Carrot', 'Teddy bear', 'Glove', 'Cowboy Hat', 'Lighthouse', 'Bowl', 'Pumpkin', 'Candle', 'Oyster', 'Handgun', 'Hedgehog', 'Scarf'],
	}%
}

\noindent\fbox{%
	\parbox{0.97\linewidth}{%
~~\textcolor{blue}{'Novel Categories'}: ['Football', 'Toy', 'Bird', 'Surfboard', 'Swimwear', 'Balloon', 'Computer keyboard', 'Shirt', 'Cheese', 'Suit', 'Cookie', 'Box', 'Drink', 'Roller skates', 'Flag', 'Guitar', 'Door handle', 'Goat', 'Bottle', 'Mobile phone', 'Skateboard', 'Raven', 'High heels', 'Train', 'Truck', 'Handbag', 'Wine', 'Wheel', 'Monkey', 'Chicken', 'Crocodile', 'Clothing', 'Lemon', 'Dog', 'Book', 'Mule', 'Jeans', 'Flowerpot', 'Parrot', 'Ball', 'Sparrow', 'Sandwich', 'Tortoise', 'Starfish', 'Suitcase', 'Cello', 'Jet ski', 'Computer mouse', 'Christmas tree', 'Dolphin', 'Rhinoceros', 'Hat', 'Kangaroo', 'Mouse', 'Plastic bag', 'Polar bear', 'Lipstick', 'Zebra', 'Eagle', 'Banana', 'Orange', 'Goose', 'Swan', ' Watch', 'Luggage and bags', 'Doughnut', 'Fire hydrant', 'Table tennis racket', 'Saxophone', 'Picture frame', 'Waste container', 'Elephant', 'Spoon', 'Ostrich', 'Harpsichord']\}
	}%
}


\section{More Implementation Details}
All the training images are resized to 256$\times$256, and then randomly cropped to 224$\times$224 for augmentation. For both baseline and our method, we use ResNet50 as the encoder and initialize it with weights pre-trained by MoCo. MLP-based projection head contained in the model is also initialized from MoCo. During inference, the images are resized to 256$\times$256 and cropped in the center to 224$\times$224.

\textbf{The proposed paradigm}. For our method, we adopt SGD optimizer with a weight decay of 0.0001 and a momentum of 0.9. For supervised contrastive learning, $N_z$ is set as 12. For semantic centroids-driven contrastive learning, we set the number of centroids as 50,000, 5,000 and 2,500 for ImageNet-1K, iNatLoc500 and OpneImages150, respectively.The number of negative samples in $\mathcal{L}_{mcl}$ is set to 16,384, 4,096 and 2,048 on ImageNet-1K, iNatLoc and OpenImages150, respectively.  The hyper-parameters $\alpha$ and $\beta$ in Equation \textcolor{red}{6} are set to 1.0 and 0.5, respectively. For ImageNet-1K dataset, we use a mini-batch size of 256 and an initial learning rate of 0.003.  For iNatLoc500 dataset, we use a mini-batch size of 128 and an initial learning rate of 0.001. For OpenImages150 dataset, we use a mini-batch size of 64 and an initial learning rate of 0.0005. Our model is implemented using PyTorch and trained on 4 NVIDIA Tesla V100 GPUs.

\textbf{Baseline from WSOL}.  For baseline methods from WSOL, we follow the implementations and learning schedules of the open-source repository\footnote{https://github.com/clovaai/wsolevaluation}. Specifically, all methods are trained for 10 epochs and the learning rate is decayed every 3 epochs using SGD optimizer. Since the parametric classification head of baselines cannot handle novel categories, we perform clustering and use the proposed generalized class activation mapping for evaluation on the \textit{test} set. Additionally, results of known categories using the parametric classification head are also given. We ensure that the re-implemented performance on known categories is close to the ones reported in the literature. 

\textbf{Baseline from NCD}. We adopt two novel categories discovery works, i.e., DTC and GCD, as the baseline of our OWSOL. Both DTC and GCD originally follow the transductive setting. For a fair comparison, we shift them to the proposed OWSOL setting. DTC first trains a classification model on the labeled data, and then takes it to initialize the encoder for transfer clustering on the unlabeled data. As categories in labeled data and unlabeled data are disjoint in the setting of DTC, we train the classification head of transfer clustering for the total number of all categories to fit the setting of OWSOL. For GCD, due to a large number of categories of  datasets, 
a mini-batch cannot uniformly incorporate samples from labeled and unlabeled data to traverse known categories, resulting in performance degradation. To mitigate this issue, we adopt the momentum encoder and a dictionary with a queue to boost the performance of GCD.

\begin{figure*}[ht]
	\centering
    \includegraphics[width=\linewidth]{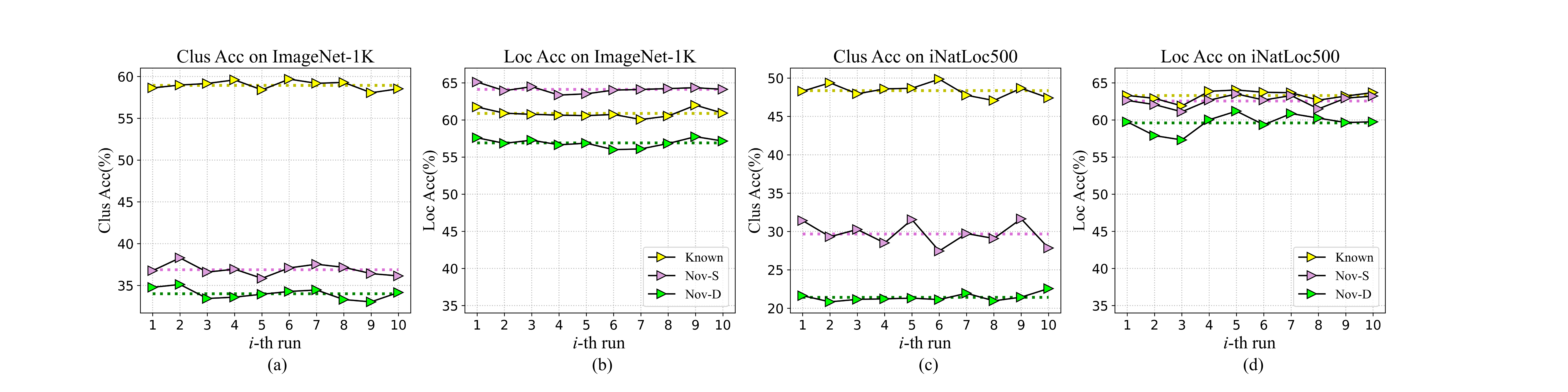}
	\caption{Results on ImageNet-1K and iNatLoc500 over 10 runs.} 
	\label{fig:robustness}
\end{figure*}

\section{Robustness Analysis} As clustering is integrated, it is necessary to investigate the robustness of the proposed paradigm. We train our method on ImageNet-1K and iNatLoc500 over 10 runs, respectively, and present the corresponding Clus Acc and  Loc Acc on ImageNet-1K and iNatLoc500 in Figure~\ref{fig:robustness} (a), (b), (c) and (d), respectively. The mean performances are illustrated as dotted lines in different colors. It is clear that, for ImageNet-1K, both Clus Acc and Loc Acc are stable during 10 runs across various categories. It means that the proposed paradigm is robust to the large-scale open-world dataset. Besides, for iNatLoc500, the Clus Acc of known and Nov-D, and Loc Acc of known and Nov-S are also relatively stable. However, there remain relatively large variances in Clus Acc of Nov-S and Loc Acc of Nov-D. As stable performances can be derived from the large-scale ImageNet-1K, we argue that the small scale of data may affect the stableness of clustering, accordingly attributing to the relatively large variances in performances. 

\section{Feature Manifold}
\begin{figure}[h]
	\centering
    \includegraphics[width=\linewidth]{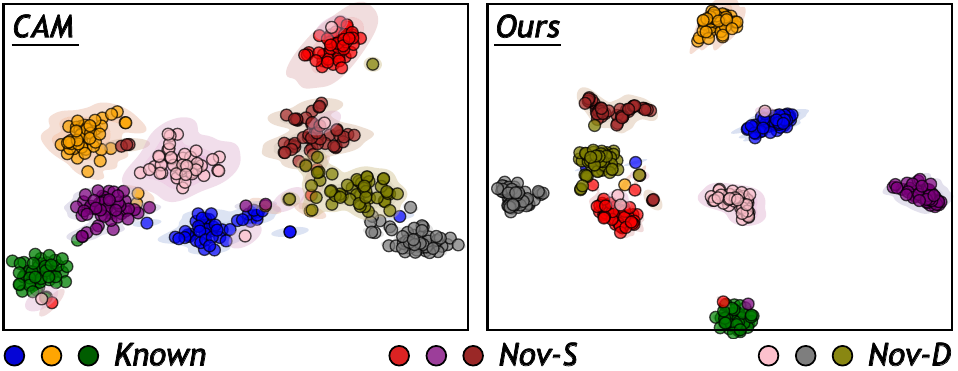}
    \caption{Feature manifold of CAM and the proposed method on the test set of ImageNet-1K. } 
	\label{fig:manifold}
\end{figure}

Figure~\ref{fig:manifold} presents the feature manifold of CAM and the proposed method. The samples are randomly chosen from the test set of ImageNet-1K. The figure shows that the features of Nov-S and Nov-D extracted by CAM are scattered across the manifold. Thus, it is challenging to discover novel categories based on these distributions. By contrast, with the proposed contrastive representation co-learning, the intra-class representations compactly distribute in the feature space and the inter-class distances are much larger. It expands a well-structured feature space for both known and novel categories, from which novel category discovery and localization can benefit.

\begin{figure*}[!ht]
	\centering
     \vspace{-25pt}
    \includegraphics[width=0.95\linewidth]{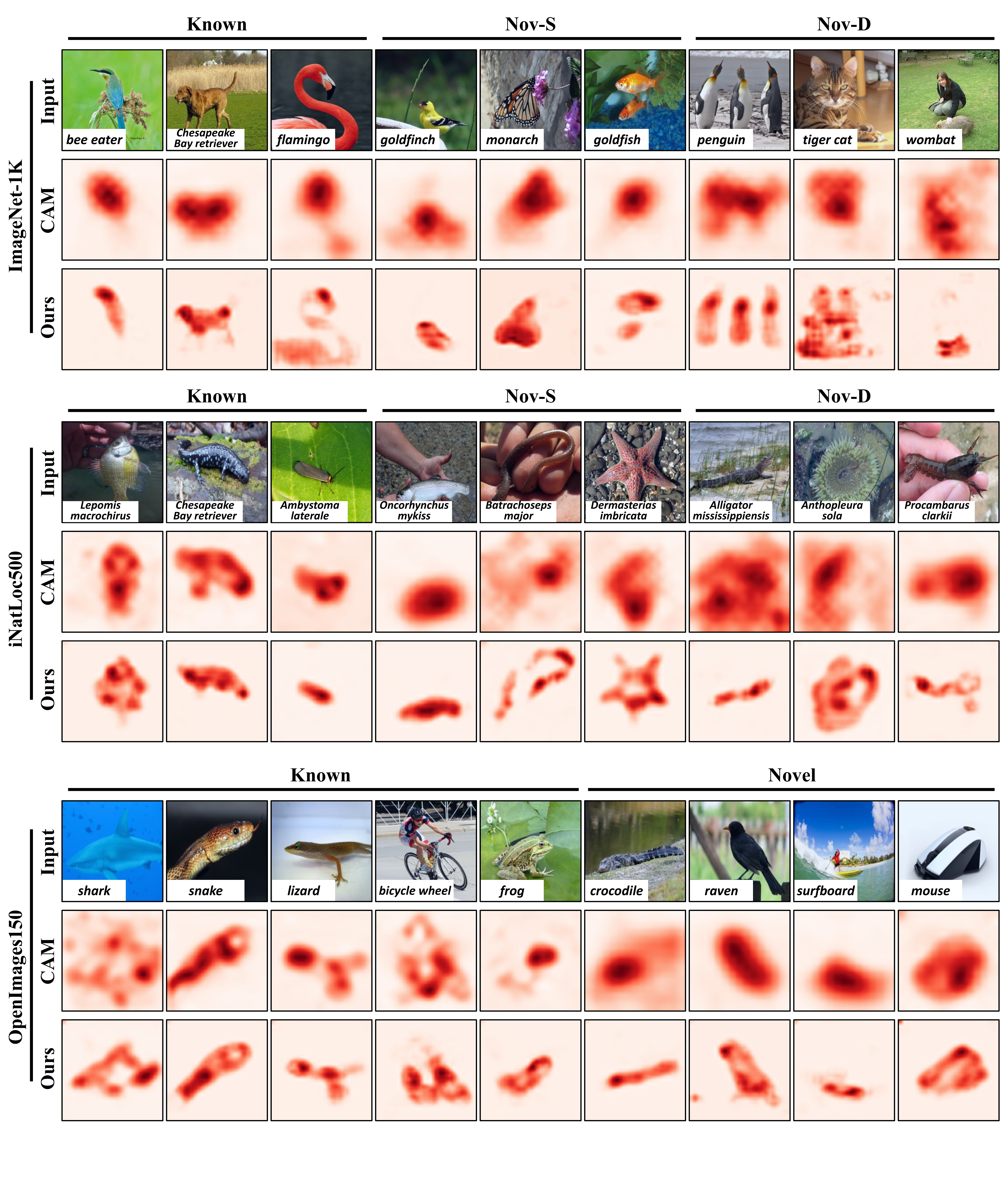}
    \vspace{-5pt}
	\caption{Visual comparisons of class activation maps on ImageNet-1K, iNatLoc500, and OpenImages150.} 
	\label{fig:visualization}
	
\end{figure*}
\section{Visual Results}

In Figure~\ref{fig:visualization}, we present the activation maps of CAM and the proposed method on ImageNet-1K, iNatLoc500, and OpenImages150, respectively. For known categories on ImageNet-1K, the salient object regions, e.g., the head regions of 'bee eater' and 'flamingo', apparently dominate the activation maps of CAM. Besides, there occasionally remains excessive diffusion of activation on novel objects like 'goldfinch' and 'penguin'. An extreme failure case is that CAM sometimes cannot discover the potential objects in the image, e.g., 'wombat' in the last column. Same phenomenons can be found in visual results on iNatLoc500 and OpenImages150 datasets. In contrast, activation maps of the proposed method can generally activate complete object regions even with fine-grained contour. For example, the full body and neck of 'flamingo' and 'penguin' in the first row can be well included in the activation maps. Further, as a novel category, 'wombat' is appropriately discovered in the activation map. In general, the activation maps of the proposed method on ImageNet-1K, iNatLoc500, and OpenImages150 include more complete activation of object regions and can well discover novel categories. This validates the effectiveness and robust transferability of the proposed method to the open world.



\section{Failure Case}
\begin{figure*}[h]
	\centering
    \includegraphics[width=0.95\linewidth]{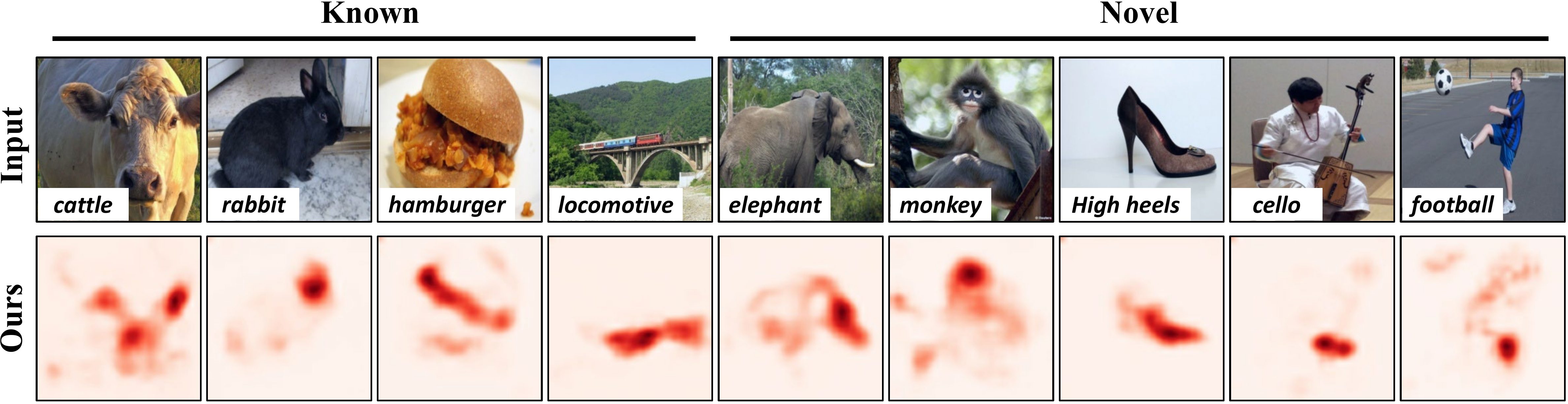}
	\caption{Failure cases of class activation maps.} 
	\label{fig:fail_case}
	\vspace{-15pt}
\end{figure*}

Figure~\ref{fig:fail_case} shows some failure cases in class activation maps of the proposed method on ImageNet-1K and OpenImages150 datasets. It can be seen that the activation maps can generally indicate the location of most objects like 'cattle', 'elephant' and 'rabbit', but sometimes focus on the most discriminative head and eye regions. Besides, for some categories like 'locomotive' and 'football' with spurious backgrounds, the model mistakenly identifies the railway and foot as target objects, which leads to the main degradation of localization accuracy.

\section{Zero-shot Ability on ImageNet-1K}
Table~\ref{tab:zero-shot} shows zero-shot ability to novel classes not seen in training for ImageNet-1K. 500 novel classes are divided into 200 and 300 novel classes called Nov-200 and Nov-300, respectively. Only Known classes and Nov-300 are used for training and the remaining Nov-200 is adopted for zero-shot evaluation. Table~\ref{tab:zero-shot} shows that our G-CAM performs much better than CAM. The Clus Acc on Nov-200 drops obviously when it's not seen in the training, while the Loc Acc is only slightly decreased from 63.76\% to 61.33\%. The results suggest that our method has a strong zero-shot ability to localize novel classes never seen before.

\begin{table}[h]
 \centering
 \caption{Zero-shot ability to classes not seen in training on ImageNet-1K.}
 \resizebox{\linewidth}{!}{
  \begin{tabular}{c c c >{\columncolor{mygray}} c c >{\columncolor{mygray}}c}
   \toprule
   \multirow{2.5}{*}{Methods} & {seen in training (\checkmark) or not ($\times$)}
   & \multicolumn{2}{c}{Clus Acc} & \multicolumn{2}{c}{Loc Acc } \\
            \cmidrule(lr){2-2} \cmidrule(lr){3-4} \cmidrule(lr){5-6}
             & Nov-200  &Nov-300 & Nov-200 &Nov-300 & Nov-200 \\ 
            \midrule
            CAM  &$\times$ &  17.78 & 19.00 & 55.11 & 57.32\\
            \rowcolor{mygray1}
            G-CAM  &$\times$ & 36.00 & 23.98 & 60.52 & 61.33  \\        
            G-CAM  &\checkmark & 34.13 & 36.86 & 61.26 & 63.76  \\ 
   \bottomrule
 \end{tabular}}
 \label{tab:zero-shot}
 \vspace{-15pt}
\end{table}



\end{document}